\documentclass{ecai}

\usepackage{graphicx}
\usepackage{latexsym}
\usepackage{booktabs}
\usepackage{microtype}
\usepackage{bm}
\usepackage{amsmath}
\usepackage{amsfonts}
\usepackage{bbm}
\usepackage{helvet}
\usepackage{courier}
\usepackage{multirow}

\usepackage{mdwlist}

\usepackage{relsize}
\usepackage{url}
\usepackage{subfigure}
\usepackage{caption}
\usepackage{hhline}
\usepackage{color}
\usepackage{colortbl}
\usepackage{booktabs}
\usepackage{tabularx}
\usepackage{amsmath}

\usepackage[linesnumbered,boxed,ruled,commentsnumbered]{algorithm2e}

\begin{document}

\begin{frontmatter}

\newcommand{\encoder}{Relation Instance Encoder}

\title{Siamese Representation Learning for Unsupervised Relation Extraction}

\author[A]{\fnms{Guangxin}~\snm{Zhang}}
\author[A]{\fnms{Shu}~\snm{Chen}\thanks{Corresponding Author. Email: kjcuse@jiangnan.edu.cn}}
\address[A]{School of IoT Engineering, Jiangnan University, China}

\begin{abstract}
Unsupervised relation extraction (URE) aims at discovering underlying relations between named entity pairs from open-domain plain text without prior information on relational distribution. Existing URE models utilizing contrastive learning, which attract positive samples and repulse negative samples to promote better separation, have got decent effect. However, fine-grained relational semantic in relationship makes spurious negative samples, damaging the inherent hierarchical structure and hindering performances. To tackle this problem, we propose Siamese Representation Learning for Unsupervised Relation Extraction -- a novel framework to simply leverage positive pairs to representation learning, possessing the capability to effectively optimize relation representation of instances and retain hierarchical information in relational feature space. Experimental results show that our model significantly advances the state-of-the-art results on two benchmark datasets and detailed analyses demonstrate the effectiveness and robustness of our proposed model on unsupervised relation extraction. We have released our code at \url{https://github.com/gxxxzhang/siamese-ure}. 
\end{abstract}

\end{frontmatter}

\section{Introduction}
%
Relation Extraction (RE) is the task of extracting semantic relation between entity pair from raw text. For example, given the sentence \textit{``ChatGPT is created by OpenAI, a research organization dedicated to creating and promoting friendly AI that benefits humanity''}, and the entity pair \textit{(ChatGPT, OpenAI)}, RE model can predict the pre-define relationship \textit{``created\_by''} and extract the corresponding triplet \textit{(ChatGPT, created\_by, OpenAI)} for downstream tasks, such as web search \cite{xiong2017explicit}, knowledge base construction \cite{al2018extracting} and question answering \cite{burns2022discovering}. Existing RE methods which are restricted to specific relation types have achieved good performance with annotated data. Nevertheless, with the rapid emergence of large, domain-specific text corpora (e.g., sports news, social media content, scientific publications) and new relation types in the real world, these methods face many challenges. On the one hand manually establishing and maintaining the ever-growing relation require expert knowledge and are time-consuming, on the other hand these methods are hard to scale up to newly emerged relations. Unsupervised relation extraction is promising and received widespread concern since it does not require prior information on relation distribution to reduce the reliance on labeled data and can discover new relation types in raw text. 

Traditional unsupervised relation extraction approaches are based on variational autoencoder (VAE) architecture \cite{marcheggiani2016discrete,simon2019unsupervised,tran-etal-2020-revisiting,yuan-eldardiry-2021-unsupervised}. These methods train the relation extraction model as an encoder that generates relation classifications. A decoder is trained along with the encoder to reconstruct the encoder input based on the encoder generated relation classifications. However, joint training for two networks (encoder and decoder) and requiring the exact number of relation classes during the training period of the encoder make model unstable. Different from treating relations as latent variables, the clustering-based approaches learn semantic relational representation from high-dimensional embeddings and adopt unsupervised clustering algorithms to recognize relation classes in feature space. In this process, the main challenge is how to learn semantic representation of instances in the relational feature space.

Elsahar et al. \cite{elsahar2017unsupervised} extracts KB types and NER tags of entities as well as re-weighted word embeddings from sentences, then adopts Principal Component Analysis (PCA) to reduce feature dimensionality that can alleviate the problem of features sparsity, and finally uses Hierarchical Agglomerative Clustering (HAC) to cluster the feature representations. Because integrating word embeddings in a rule-based way, the method heavily rely on hand-craft features and make many simplifying assumptions and its feature space is lack of semantic information. Liu et al. \cite{liu-etal-2021-element} formulate URE using a structural causal model and conduct Element Intervention to eliminate spurious correlations, which intervenes on the context and entities respectively to obtain the underlying causal effects of them and learn the causal effects through instance-wise contrastive learning. 

\begin{figure*}[t!]
    \centering
    \includegraphics[width=1\linewidth]{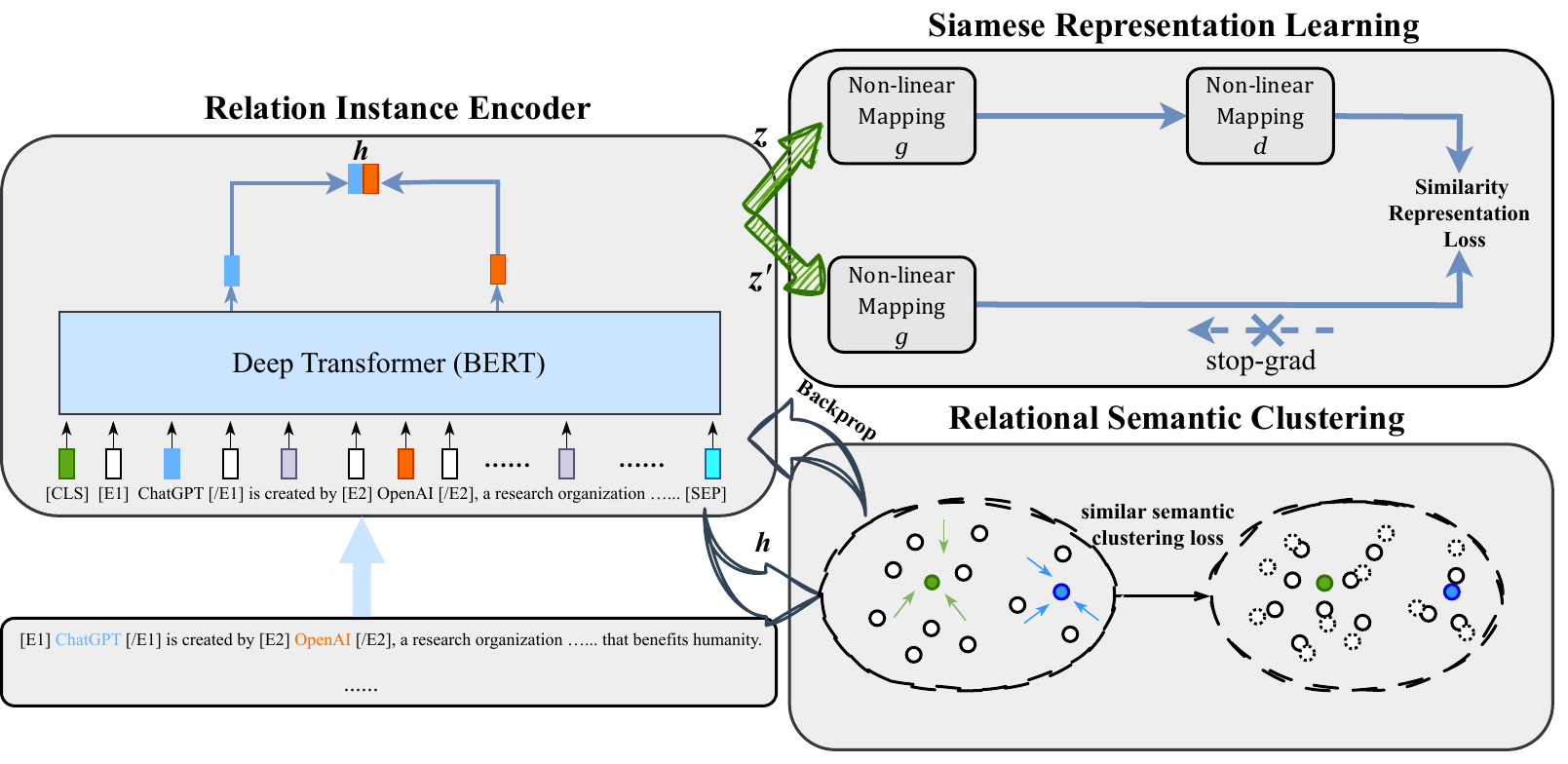}
    \caption{Overall architecture of our model. Instances will be fed into encoder with different dropout masks $z$ and $z^{\prime}$ to obtain feature pairs, then transmitted into Non-linear Mapping $\bm{g}(\cdot)$ and $\bm{d}(\cdot)$ respectively. We use ones to predict other ones to optimize encoder. Besides, in relational semantic clustering, we mining nearest neighbors of each sample with semantic clustering loss.}
    \label{figure:model:overview}
\end{figure*}

The core idea of instance-wise contrastive learning is to pull together the representations within positive instances while pushing apart negative ones. However, negative examples are commonly sampled from the batch or training data at random due to the lack of ground-truth annotations. In relation extraction, the relational semantic tend to be more fine-grained and based on a potential hierarchical structure \cite{zhang-etal-2021-open}. For example, the relations \textit{``per:stateorprovinces\_of\_residence''}, \textit{``per:countries\_of\_residence''} and \textit{``per:cities\_of\_residence''} in one benchmark dataset share the same parent semantic on \textit{/people/residence}, which means that they belong to the same semantic cluster from a hierarchical perspective. Naturally, instances may have highly similar semantics in a batch but contrastive learning pushes these representations apart as long as they are from different original instances, regardless of their semantic similarities. To alleviate the dilemma which instance-wise contrastive learning unreasonably pushes apart those sentence pairs that are semantically similar, Liu et al. \cite{liu-etal-2022-hiure} propose hierarchical exemplar contrastive learning. The model leverages Hierarchical Propagation Clustering to obtain hierarchical exemplars from relational feature space and further utilizes exemplars to hierarchically update relational features of sentences and is optimized by performing both instance and exemplar-wise contrastive learning through Hierarchical Exemplar Contrastive Loss and propagation clustering iteratively.  Nonetheless, the method still utilize traditional instance-wise contrastive learning loss to retain the local smoothness in relational feature space during training period and use an iterative way to obtain hierarchical exemplars will cause error accumulation \cite{palanivinayagam2020optimized}.

In this paper, we attempt to propose a Siamese network architecture, only using positive pairs for representation learning, to eliminate adverse effect of spurious negative samples. Siamese networks are able to learn similarity metrics of relations from labeled data of pre-defined relations, and then transfer the relational knowledge to identify novel relations in unlabeled data \cite{wu-etal-2019-open}. Therefore, we intend to use Siamese architecture to construct positive pairs and learn relation representations in unsupervised setting. For the most part, owing to lack of labeled data, the network is easy to collapse (i.e., all outputs “collapsing” to a constant). To avoid that ,we introduce entity type, which provide a strong inductive bias for relation extraction \cite{tran-etal-2020-revisiting}, as prior information. Furthermore, followed by Chen and He \cite{Chen_2021_CVPR}, we use the analogous network architecture to further prevent to collapse. To recognize different relations, we through traditional unsupervised clustering algorithms (e.g., k-means clustering algorithm) to cluster learned representation in relational feature space. Nevertheless, naively applying these clustering algorithms on the obtained features can lead to cluster degeneracy \cite{caron2018deep}. We propose relational semantic clustering module that mining nearest neighbors of each instance in feature space. The module can support model learn more discriminative representations under semantic clustering loss, so that the learned representation is cluster-friendly and obtain better clustering performance.

Our main contributions are the following: (1) We propose a novel representation learning framework for unsupervised relation extraction. Furthermore, our model is much simpler than existing self-supervised learning models that apply empirical data augmentation and complicated network architecture. (2) We explore Relational Semantic Clustering module, encoding the relational semantic into the representations via unsupervised clustering. We efficiently conduct representation learning and unsupervised clustering in a unified framework. (3) We conduct extensive experiments on two datasets and achieves better performance than the existing state-of-the-art methods. Meanwhile, our ablation analysis shows the impacts of different modules in our framework.

\section{Model}
We aim at developing a joint model that leverages the beneficial properties of self-supervised learning to improve unsupervised relation extraction. As illustrated in Figure \ref{figure:model:overview}, our model consists of three modules: Relation Instance Encoder, Siamese Representation Learning and Relational Semantic Clustering. The encoder module uses instances as input which are composed of natural language sentences and entity pairs, and then employs the pre-trained model to output entity-level feature pair sets $\bm{H}$ and $\bm{H^{\prime}}$ for all instances. The learning module is structured as Siamese architecture and takes pair sets respectively as input. We use similarity representation loss, which measure the cosine similarity of positive pairs, to enforces the relational feature of instances that have similar semantics to be more close in feature space. In clustering module, we mining nearest neighbors of each sample in relational feature space to learn discriminative representations under semantic clustering loss that can yield better clustering performance.

\subsection{Relation Instance Encoder}
The Relation Instance Encoder aims to obtain relational features from instances. We use instances $\bm{X}$ as inputs that each $\bm{x_i}\in\bm{X}$ is composed of a sentence $s_i={\{w_{1}, ..,w_{n}\}}$ with $n$ words, $\operatorname{h_i}=(\operatorname{h_i^s},\operatorname{h_i^e})$ representing the start and end position of head entity, and $\operatorname{t_i}=(\operatorname{t_i^s},\operatorname{t_i^e})$ representing the start or end position of tail entity in the sentence. Same as the previous work, named entities in the sentences have been recognized in advance. We employ pre-trained BERT \cite{devlin-etal-2019-bert} model, which has strong performance on extracting contextual information, as encoder $\bm{f}$ to map relation instance $x_i$ to embedding. However, BERT always induces a non-smooth anisotropic semantic space of sentences \cite{li2020sentence} , which is easier make model to collapse. To this end, we add entity types as prior information in head and tail entities  as $[\operatorname{h_i^s}]$, $[\operatorname{h_i^e}]$, $[\operatorname{t_i^s}]$, $[\operatorname{t_i^e}]$ and inject them to each instance $\bm{x_i}$:
\begin{equation}
\bm{x_i}=[{w_1,\ldots,[\operatorname{h_i^s}],\ldots,w_i,\ldots,[\operatorname{h_i^e}],\ldots,[\operatorname{t_i^s}],\ldots,[\operatorname{t_i^e}],\ldots,w_n}]
\end{equation}
then get the token embedding:
\begin{equation}
    \mathbf{e}_{1},...,\mathbf{e}_{n}=\bm{f}_{\theta}\mathbf{(x_{1},...,x_{n})}
\end{equation}
where ${\theta}$ is the learnable parameters in the encoder. We use the embeddings with position of $[\operatorname{h_i^s}]$ and position of $[\operatorname{t_i^s}]$ as outputs to obtain the entity-level feature ${\bm{h}_{i}\in2\cdot{\mathbb{R}^d}}$:
\begin{equation}
\bm{h}_{i}=\mathbf{e}_{head}\oplus\mathbf{e}_{tail}
\end{equation}
We follow the \textbf{data augmentation} used in SimCSE \cite{gao-etal-2021-simcse} to construct positive pairs. Specifically, we only feed the same input $\bm{x_i}$ to the encoder twice with different dropout masks, which  placed on fully-connected layers as well as attention probabilities, and we can obtain two different embeddings as positive pairs. We denote positive pair as $\bm{h_i}=\bm{f}_{\theta}\left(\mathbf{x_i};\ \mathbf{z}\right)$ and $\bm{h_i^\prime}=\bm{f}_{\theta}\left(\mathbf{x_i};\ \mathbf{z^\prime}\right)$, where $\mathbf{z}$ and $\mathbf{z^\prime}$ is the different dropout masks.
\subsection{Siamese Representation Learning}
We use Siamese network which can naturally introduce inductive biases for modeling invariance to learn relational similarity metrics. However, Siamese networks suffers from the problem of model collapse, where the model converges to a constant value and the samples all mapped to a single point. Besides, it is difficult to learn a reasonable distance in feature space without negative samples. Followed by Chen and He \cite{Chen_2021_CVPR}, we attempt to use the similar approach to address this issue. As shown in Figure \ref{figure:model:overview}, 
positive feature pair $\bm{h_i}$ and $\bm{h_i^\prime}$ of one instance are processed by the same encoder network $\bm{f}$ with different dropout masks $\mathbf{z}$ and $\mathbf{z^\prime}$. Then we use the MLP $\bm{g}(\cdot)$ to map feature pair to ${g_i}$ and ${g_i^\prime}$ respectively: 
\begin{equation}
\label{map:}
    {g_i},{g_i^\prime}=\bm{g}_{\phi}\left(\bm{h_i},\ \bm{h_i^\prime}\right)
\end{equation}
where ${\phi}$ is the learnable parameters in the non-linear mapping network. And then, the non-linear mapping network $\bm{d}(\cdot)$ is applied on one side, and we denote the feature as ${d_i^\prime}=\bm{d}_{\psi}({g_i^\prime})$, ${\psi}$ is the learnable parameters. Meanwhile, a stop-gradient operation is applied on the other side. The model maximizes the similarity between both sides to learn relation representation under similarity representation loss. 
\noindent\textbf{Similarity Representation Loss}\\
Given a training set $\bm{X}=\{{\bm{x_{1}},\bm{x_{2}},\dots,\bm{x_{n}}}\}$ of $n$ instances, Relation Instance Encoder can obtain two augmented relational features for each input sentences by feed the same input to the encoder twice with different dropout masks. In this process, we obtain feature sets $\bm{H}=\{{\bm{h_{1}},\bm{h_{2}},\dots,\bm{h_{n}}}\}$ and $\bm{H^\prime}=\{{\bm{h^\prime_{1}},\bm{h^\prime_{2}},\dots,\bm{h^\prime_{n}}}\}$. 
We minimize negative cosine similarity between positive pair:
\begin{equation}
    \mathcal{L}_{Rl}=\frac{1}{2}\left[\mathcal{D}\big(d,\text{stopgrad}(g^\prime)\big)+\mathcal{D}\big(d^\prime,\text{stopgrad}(g)\big)\right]
\end{equation}
where $\text{stopgrad}(\cdot)$ is use stop-grad strategy that gradient does not back-propagate. The cost is described as a symmetrized form since pairs from the Siamese network. For each part of the loss, we minimize their negative cosine similarity:
\begin{equation}
    \mathcal{D}(d,g^\prime)=-\frac{1}{n}\sum_{i=1}^n\frac{d_i}{\|d_i\|_2}\cdot\frac{g^\prime_i}{\|g^\prime_i\|_2}
\end{equation}
for a mini-batch of N instances,where $i\in{[1,N]}$ and ${\left\lVert{\cdot}\right\rVert _2}$ is $\ell_2$-norm.

\subsection{Relational Semantic Clustering}
One of the main obstacles for our model is difficult learn a discriminative representation without negative samples. Be enlightened by Van Gansbeke \cite{van2020scan}, we assume that in a excellent relational feature space, each sample with their nearest neighbors have similar relational semantic and belong to the same relation class. We propose Relational Semantic Clustering to learn discriminative relational representations and conduct clustering and representation learning in a unified framework. Specifically, for each instance $\mathbf{x_{i}}$, we mine its K nearest neighbors according to each representation $\bm{h_{i}}$ and define the set $\mathcal{N}_{x_{i}}$ as the output features of neighboring samples corresponding to each $\mathbf{x_{i}}$ in the dataset. 
Then, we use similar semantic clustering loss to attract instances and  their neighboring samples to approach each other. Simultaneously, different instances are separated in feature space.

\noindent\textbf{Similar Semantic Clustering Loss}\\
Like adaptive clustering \cite{xie2016unsupervised}, we aim to learn a clustering function $\Phi_\sigma$ - parameterized by a neural network. The neural network classifies each instance $\mathbf{x_i}$ and its mined neighbors $\mathcal{N}_{x_i}$ together with the learnable parameters $\sigma$.
 The function $\Phi_\sigma$ terminates in a softmax function to perform a soft assignment over the clusters $\mathcal{C}=\left\{1,\ldots,C\right\}$, with $\Phi_\sigma \left(\mathbf{x_i}\right) \in [0,1]^C$. The probability of instance $\mathbf{x_i}$ being assigned to cluster $c$ is denoted as $\Phi_\sigma^c\mathbf{(x_i)}$. We learn the weights of $\Phi_\sigma$ by minimizing the following objective:
\begin{equation}
\label{eq:loss_objective}
\mathcal{L}_{Cl} = -\frac{1}{|\mathbf{      X}|}\sum\limits_{\mathbf{x_i}\in\mathbf{X}}\sum\limits_{k\in\mathcal{N}_{\mathbf{x_i}}}\log\left<\Phi_\sigma(\mathbf{x_i}),\Phi_\sigma(k)\right> + \lambda\sum_{c\in\mathcal{C}} \Phi_\sigma'^c \log \Phi_\sigma'^c 
\end{equation}
where $\left<\cdot\right>$ denotes the dot product operator. 
The first term in Equation~\ref{eq:loss_objective} imposes $\Phi_\sigma$ to make consistent predictions for each instance $\mathbf{x_i}$ and its neighboring samples $\mathcal{N}_{\mathbf{x_i}}$. However, the dot product will be maximal when the predictions are one-hot (confident) and assigned to the same cluster (consistent). To avoid $\Phi_\sigma$ from assigning all samples to a single cluster, we include an entropy term(the second term in Equation~\ref{eq:loss_objective}):
\begin{equation}
\Phi_\sigma'^c = \frac{1}{|\mathcal{\mathbf{X}}|}\sum\limits_{\mathbf{x_i}\in\mathcal{\mathbf{X}}}\Phi_\sigma^c(\mathbf{x_i}).
\end{equation}
which spreads the predictions uniformly across the clusters $\mathcal{C}$ and encourages the classifier to scatter a set of instances into different classes.
\subsection{Iterative Joint Training}
In early training period, we only optimize $\mathcal{L}_{Rl}$ in Siamese Representation Learning module to drive instances with similar semantic get closer in feature space. After several warm-up epochs, instances have acquired reasonable semantic representations. We introduce Relational Semantic Clustering module to further refine the semantic representations and enable them to be more discriminative.  
In summary, our overall objective is:
\begin{equation}
    \mathcal{L}=\mathcal{L}_{Rl}+\eta\mathcal{L}_{Cl}
\end{equation}
where $\eta$ is a loss coefficient.
Our approach involves the combined utilization of the Siamese Representation Learning module and the Relational Semantic Clustering module through an iterative procedure. This joint usage enables model to achieve a well-separated representation of distinct instances in the learned feature space while preserving local invariance for each individual instance.

\section{Experiments}
In this section, we first describe two relation extraction datasets for training and evaluating the proposed method, then detail the baseline models for comparison, and then expound the implementation details and hyperparameter configuration, finally we conduct a comprehensive and detailed analysis of our model.
\begin{figure}[t]
    \centering
    \includegraphics[width=1.\columnwidth]{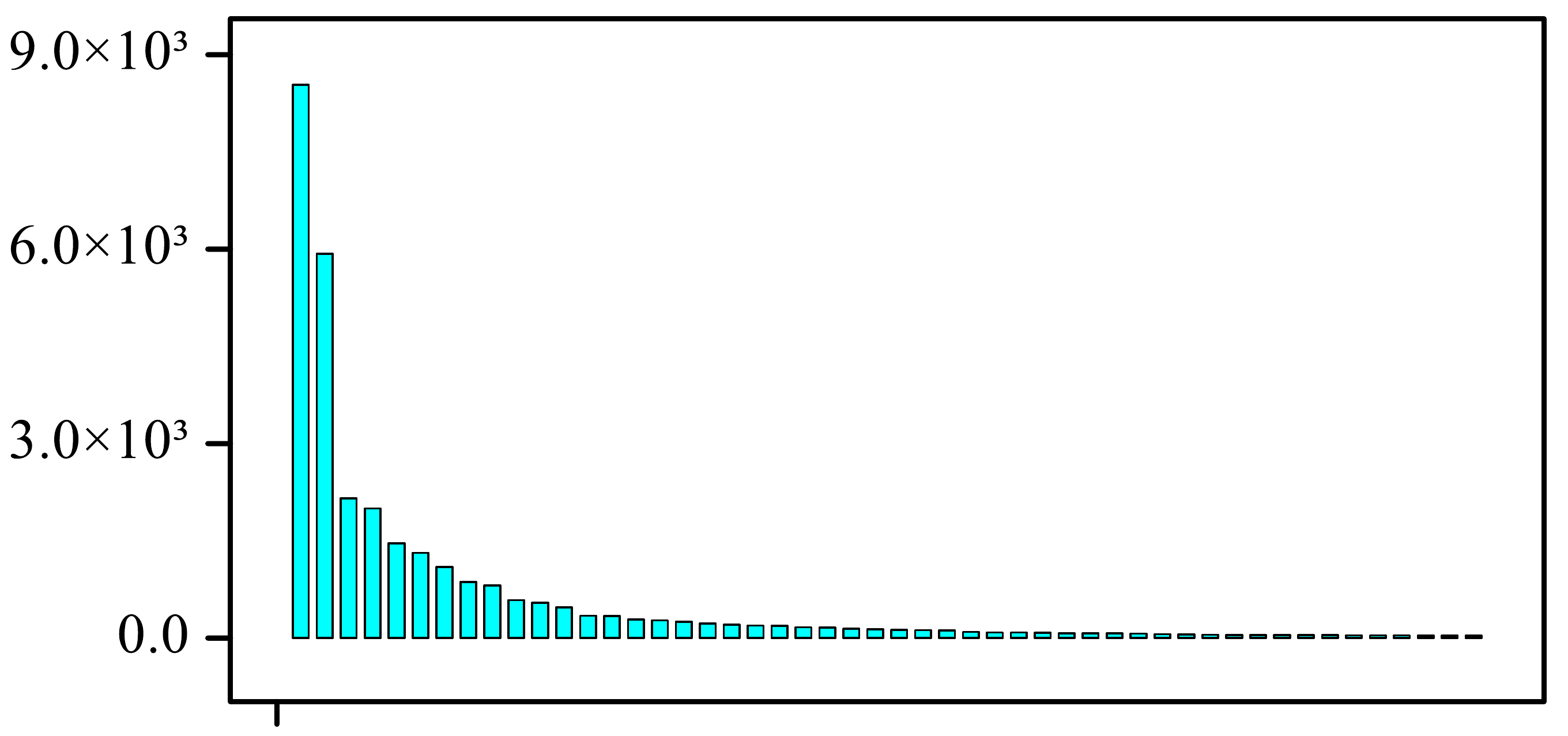}
    \caption{The relation distribution in the NYT+FB dataset. The ordinate represents the number of sentences in each relation type in the dataset.The x-axis is the relations sorted according to the number of sentences contained. For ease of observation, the x-axis label is omitted.}
    \label{fig:dataset distribution}
\end{figure}

\subsection{Datasets and Evaluation Metrics}
\noindent\textbf{Datasets.}\quad 
Following previous work \cite{tran-etal-2020-revisiting,liu-etal-2022-hiure}, we conduct experiments on two relation extraction datasets -- NYT+FB \cite{marcheggiani2016discrete} and TACRED \cite{zhang-etal-2017-position} with different constructing settings. The former is generated via distant supervision while the latter is manually annotated corpus, which is extremely challenging to model. The NYT+FB dataset is obtained by using Freebase to label the corpus of the New York Times corpus. That is, if the entity pair that appears in a sentence also appears in Freebase
\cite{bollacker2008freebase}, then this sentence is automatically labeled as the relation stored by Freebase. After filtering out some sentences using syntactic patterns, there are 2 million sentences in the dataset, of which 41,000 are labeled with meaningful relations1. Of the 41,000 tagged sentences, 20\% are used as validation set, and 80\% are used as test set. The TACRED dataset is a large-scale crowd-sourced relation extraction dataset following the TAC KBP relation schema that covers 42 relation types. We remove the instances labeled as \texttt{no\_relation} and use the remaining 21,773 instances including 41 relation types for training and evaluation.

\noindent\textbf{Evaluation Metrics.}\quad 
B-cube (${\rm B}^3$) \cite{bagga-baldwin-1998-entity}, V-measure \cite{rosenberg-hirschberg-2007-v} and Adjusted Rand Index (ARI) \cite{hubert1985comparing} are used as evaluation metrics for different models. Specifically, ${\rm B}^3$ contains the precision and recall metrics to correspondingly measure the correct rate of putting each sentence in its cluster or clustering all samples into a single class, which are defined as follows:
\begin{equation*}
  \setlength{\abovedisplayskip}{3pt}
  \setlength{\belowdisplayskip}{3pt}
  \begin{aligned}
    \rm B^3_{Prec.}=& \mathop{\mathbb{E}}\limits_{X,Y}P(g(X)=g(Y)|c(X)=c(Y))\\
    \rm B^3_{Rec.}=& \mathop{\mathbb{E}}\limits_{X,Y}P(c(X)=c(Y)|g(X)=g(Y))\\
  \end{aligned}
\end{equation*}
Then ${\rm B}^3$\ \ $\rm F_1$ is computed as the harmonic mean of the precision and recall. 

V-measures contains the homogeneity and completeness, which is analogous to ${\rm B}^3$ precision and recall. These two metrics penalize small impurities in a relatively ``pure'' cluster more harshly than in less ``pure'' ones:
\begin{equation*}
  \setlength{\abovedisplayskip}{3pt}
  \setlength{\belowdisplayskip}{3pt}
  \begin{aligned}
    \rm V_{Homo.}=& 1- H(c(X)|g(X))/H(c(X))\\
    \rm V_{Comp.}=& 1-H(g(X)|c(X))/H(g(x))\\
  \end{aligned}
\end{equation*}

ARI is a normalization of the Rand Index, which measures the agreement degree between the cluster and golden distribution. This metric ranges in [-1,1]. The larger the value, the more consistent the clustering result
is with the real situation.
\subsection{Baselines}
To evaluate the effectiveness of our method, we select the following unsupervised relation extraction models for comparison with standard evaluation metrics: 1) \textbf{rel-LDA} \cite{yao2011structured}, a generative model that considers the unsupervised relation extraction as a topic model.We choose the full rel-LDA with a total number of 8 features for comparison in our experiment. 2) \textbf{March} \cite{marcheggiani2016discrete}, a VAE-based model learned by self-supervised signal of entity link predictor. 3) \textbf{UIE} \cite{simon2019unsupervised},a discriminative model that adopts additional regularization to guide model learning. And it has different versions according to the choices of different relation encoding models (e.g., PCNN). We report the results of two versions—UIE-PCNN and UIE-BERT (i.e., using PCNN and BERT as the relation encoding models) with the highest performance. 4) \textbf{EType} \cite{tran-etal-2020-revisiting}, a simple and effective method relying only on entity types. The same link predictor as in \textbf{March} \cite{marcheggiani2016discrete} is employed and two additional regularizers are used. 5) \textbf{SelfORE} \cite{hu2020selfore},a self-supervised framework that bootstraps to learn a contextual relation representation through adaptive clustering and pseudo label. 6) \textbf{EIURE} \cite{liu-etal-2021-element}, a contrastive learning framework that intervenes on the context and entities respectively to obtain the underlying causal effects of them. 7) \textbf{HiURE} \cite{liu-etal-2022-hiure}, is the state-of-the-art method that derive hierarchical signals from relational feature space using cross hierarchy attention and effectively optimize relation representation of sentences under exemplar-wise contrastive learning.
\begin{table}[t]
    \renewcommand\arraystretch{1.2}
    \caption{Hyper-parameter values used in our experiments.}
    \label{tab:hyper}
    \centering
    \begin{tabular}{l c r}
    \toprule
    \textbf{Hyper-parameters} &     &\textbf{value}\\
    \midrule
    optimizer & &\textit{SGD}\\
    learning rate & &1e-5\\
    weight\_decay & &1e-4 \\
    momentum & &0.9\\
    batch size & &64\\
    warm-up epochs $L$ & &5\\ 
    dropout rate $r$ & &0.1\\
    number of nearest neighbors $K$ & &20\\
    loss coefficient $\eta$ & &0.5\\
    \bottomrule
    \end{tabular}
\end{table}
\begin{table*}
\caption{Main results on two relation extraction datasets.The results of all baseline are reproduced in liu et al. \cite{liu-etal-2022-hiure}, MLPs refers to two mapping networks in Representation Learnin module and Semantic Clustering refers to Relational Semantic Cluster module in our model.}\label{tab:result}
\centering
  \resizebox{0.95\linewidth}{!}{
\begin{tabular}{clccccccc}
\bottomrule
\multirow{2}{*}{\textbf{Dataset}} & \multirow{2}{*}{\textbf{Model}}       & \multicolumn{3}{c}{\textbf{$\text{B}^{3}$}}                       & \multicolumn{3}{c}{\textbf{V-measure}}                                                           & \multirow{2}{*}{\textbf{ARI}}               \\ \cmidrule(lr){3-5}\cmidrule(lr){6-8}
                        &&F1            & Prec.         & Rec.          & F1            & Hom.          & Comp.         &                                   \\ \hline
\multirow{11}{*}{NYT+FB}  & {rel-LDA\cite{yao2011structured}}     & 29.1\scriptsize±2.5          & 24.8\scriptsize±3.2          & 35.2\scriptsize±2.1          & \multicolumn{1}{c}{30.0\scriptsize±2.3}          &     \multicolumn{1}{c}{26.1\scriptsize±3.3}          & 35.1\scriptsize±3.5          & \multicolumn{1}{c}{13.3\scriptsize±2.7}          \\ 
                         & March\cite{marcheggiani2016discrete}            & 35.2\scriptsize±3.5          & 23.8\scriptsize±3.2         & 67.1\scriptsize±4.1 & 27.0\scriptsize±3.0         & 18.6 \scriptsize±1.8         & 49.6\scriptsize±3.1         & \multicolumn{1}{c}{18.7\scriptsize±2.6}          \\ 
                         & UIE-PCNN\cite{simon2019unsupervised}              & 37.5\scriptsize±2.9          & 31.1\scriptsize±3.0         & 47.4\scriptsize±2.8        & 38.7\scriptsize±3.2         & 32.6\scriptsize±3.3          & 47.8\scriptsize±2.9          & \multicolumn{1}{c}{27.6\scriptsize±2.5}          \\ 
                         & UIE-BERT\cite{simon2019unsupervised} & 38.7\scriptsize±2.8         & 32.2\scriptsize±2.4         & 48.5\scriptsize±2.9         & 37.8\scriptsize±2.1         & 32.3\scriptsize±2.9          & 45.7\scriptsize±3.1         & \multicolumn{1}{c}{29.4\scriptsize±2.3}          \\ 
                         & EType\cite{tran-etal-2020-revisiting} 
                         & 41.9\scriptsize±2.0          & 31.3\scriptsize±2.1          & 63.7\scriptsize±2.0          & 40.6\scriptsize±2.2          & 31.8\scriptsize±2.5          & 56.2\scriptsize±1.8         & \multicolumn{1}{c}{32.7\scriptsize±1.9}          \\  
                         & SelfORE\cite{hu2020selfore} & 41.4\scriptsize±1.9         & 38.5\scriptsize±2.2          & 44.7\scriptsize±1.8         & 40.4\scriptsize±1.7          & 37.8\scriptsize±2.4        & 43.3\scriptsize±1.9          & \multicolumn{1}{c}{35.0\scriptsize±2.0}   \\     
                         & EIURE\cite{liu-etal-2021-element} 
                         & 43.1\scriptsize±1.8 	& 48.4\scriptsize±1.9 	& 38.8\scriptsize±1.8 	& 42.7\scriptsize±1.6 	& 37.7\scriptsize±1.5 	& 49.2\scriptsize±1.6  	& \multicolumn{1}{c}{34.5\scriptsize±1.4} 
                         \\ 
                         & HiURE\cite{liu-etal-2022-hiure} 
                         & 44.3\scriptsize±0.5 	& 39.9\scriptsize±0.6 	& 49.8\scriptsize±0.5 	& 44.9\scriptsize±0.4 	& 40.0\scriptsize±0.5 	& 51.2\scriptsize±0.4  	& \multicolumn{1}{c}{38.3\scriptsize±0.6} 
                         \\ 
                        & {Our} w/o MLP\scriptsize{s}
                         & 40.5\scriptsize±0.6         & 35.9\scriptsize±0.5          & 46.5\scriptsize±0.8          & 43.3\scriptsize±0.4          & 38.2\scriptsize±0.2          & 50.0\scriptsize±0.5          & \multicolumn{1}{c}{31.0\scriptsize±0.5}          \\  
                         & {Our} w/o Semantic Clustering
                         & 41.9\scriptsize±0.3         & 36.8\scriptsize±0.3          & 48.8\scriptsize±0.4          & 44.7\scriptsize±0.8         & 39.3\scriptsize±0.2          & 51.8\scriptsize±0.9          & \multicolumn{1}{c}{32.1\scriptsize±0.7}          \\ 
                         & {Our} & \textbf{44.9\scriptsize±0.4} & 39.5\scriptsize±0.3 & 52.1\scriptsize±0.7         & \textbf{45.7\scriptsize±0.6} & 40.0\scriptsize±0.3 & 53.2\scriptsize±0.8         &
                         \multicolumn{1}{c}{\textbf{39.6\scriptsize±0.3}} \\
                         \hline \hline
                
\multirow{9}{*}{TACRED}  
                        & rel-LDA\cite{yao2011structured}          
                        & 35.6\scriptsize±2.6 	& 32.9\scriptsize±2.5 	& 38.8\scriptsize±3.1 	& 38.0\scriptsize±3.5 	& 33.7\scriptsize±2.6 	& 43.6\scriptsize±3.7 & 
                        \multicolumn{1}{c}{21.9\scriptsize±2.6}          \\  
                         & March\cite{marcheggiani2016discrete}         
                         & 38.8\scriptsize±2.9 	& 35.5\scriptsize±2.8 	& 42.7\scriptsize±3.2 	& 40.6\scriptsize±3.1 	& 36.1\scriptsize±2.7 	& 46.5\scriptsize±3.2 & 
                         \multicolumn{1}{c}{25.3\scriptsize±2.7}          \\  
                         & UIE-PCNN\cite{simon2019unsupervised} 
                         & 41.4\scriptsize±2.4 	& 44.0\scriptsize±2.7 	& 39.1\scriptsize±2.1 	& 41.3\scriptsize±2.3 	& 40.6\scriptsize±2.2 	& 42.1\scriptsize±2.6 & 
                         \multicolumn{1}{c}{30.6\scriptsize±2.5}          \\ 
                         & UIE-BERT\cite{simon2019unsupervised}  
                         & 43.1\scriptsize±2.0 	& 43.1\scriptsize±1.9 	& 43.2\scriptsize±2.3 	& 49.4\scriptsize±2.1 	& 48.8\scriptsize±2.1 	& 50.1\scriptsize±2.5 &
                         \multicolumn{1}{c}{32.5\scriptsize±2.4}          \\
                         & EType\cite{tran-etal-2020-revisiting} 
                         & 49.3\scriptsize±1.9 	& 51.9\scriptsize±2.1 	& 47.0\scriptsize±1.8 	& 53.6\scriptsize±2.2 	& 52.5\scriptsize±2.1 	& 54.8\scriptsize±1.9 & 
                         \multicolumn{1}{c}{35.7\scriptsize±2.1}          \\ 
                         & SelfORE\cite{hu2020selfore} 
                         & 47.6\scriptsize±1.7 	& 51.6\scriptsize±2.0 	& 44.2\scriptsize±1.9 	& 52.1\scriptsize±2.2 	& 51.3\scriptsize±2.0 	& 52.9\scriptsize±2.3 & 
                         \multicolumn{1}{c}{36.1\scriptsize±2.0}          \\ 
                         & EIURE\cite{liu-etal-2021-element} 
                         & 52.2\scriptsize±1.4 	& 57.4\scriptsize±1.3 	& 47.8\scriptsize±1.5 	& 58.7\scriptsize±1.2 	& 57.7\scriptsize±1.4 	& 59.7\scriptsize±1.7 &
                         \multicolumn{1}{c}{38.6\scriptsize±1.1}
                         \\ 
                         & HiURE\cite{liu-etal-2022-hiure} 
                         & 55.8\scriptsize±0.4 	& 57.8\scriptsize±0.3 	& 54.0\scriptsize±0.5 	& 59.7\scriptsize±0.6 	& 57.6\scriptsize±0.5 	& 61.9\scriptsize±0.6 & \multicolumn{1}{c}{40.5\scriptsize±0.4} 
                         \\ 
                         & {Our} w/o MLP\scriptsize{s}
                         & 53.6\scriptsize±0.8 	& 45.6\scriptsize±0.6 	& 65.0\scriptsize±1.2 	& 59.5\scriptsize±0.7          & 51.9\scriptsize±0.5          & 69.8\scriptsize±1.1          & \multicolumn{1}{c}{44.0\scriptsize±0.9}          \\ 
                        & {Our} w/o Semantic Clustering
                          & 56.1\scriptsize±0.3          & 47.7\scriptsize±0.2          & 68.1\scriptsize±0.8          & 64.1\scriptsize±0.7          & 56.2\scriptsize±0.8          & 74.6\scriptsize±1.3          & \multicolumn{1}{c}{48.4\scriptsize±0.6}          \\ 
                        & {Our} & \textbf{59.5\scriptsize±0.6} & 49.4\scriptsize±0.4 & 74.9\scriptsize±0.8        & \textbf{66.7\scriptsize±0.8} & 58.1\scriptsize±0.6 & 78.5\scriptsize±0.9        & \multicolumn{1}{c}{\textbf{50.6\scriptsize±0.5}} \\ \bottomrule
\end{tabular}}
\end{table*}
\subsection{Implementation Details}
In order to do a fair comparison with baseline method, we adopted the setting by clustering all samples into 10 relation super-classes. 
In the process of training of our model, we used the development set to manually search part of the hyper-parameters, Table \ref{tab:hyper} shows our best parameter settings. In our implementation, we adopt the pre-trained \texttt{Bert-Base-Cased} model to initialize parameters for Relation Instance Encoder and set dropout rate $r=0.1$ to generate positive pairs. The output entity-level features $\mathbf{h}_{i}$ and $\mathbf{h}_{i}^{\prime}$ possess the dimension of ${2\cdot{\mathbb{R}^d}}$, where $\mathbb{R}^d=768$. For Siamese Representation Learning, we use Non-linear Mapping $\bm{g}(\cdot)$ and $\bm{d}(\cdot)$ in our network and use SGD with 1e-5 learning rate to optimize the loss. The $\bm{g}(\cdot)$ has layer normalization (LN) \cite{ba2016layer} applied to each fully-connected (fc) layer, including its output fc. Its output fc has no ReLU. This MLP has 3 layers. The $\bm{d}(\cdot)$ has LN applied to its hidden fc layers. Its output fc does not have LN or ReLU. This MLP has 2 layers. For Relation Semantic Clustering, we set warm-up epochs $L=5$ and number of nearest neighbors of each instance $K=20$. In the evaluation period, we simply adopt the pre-trained models for representation extraction, then cluster the evaluation instances based on these representations.

\begin{figure}[t]
    \centering
    \includegraphics[width=1\linewidth]{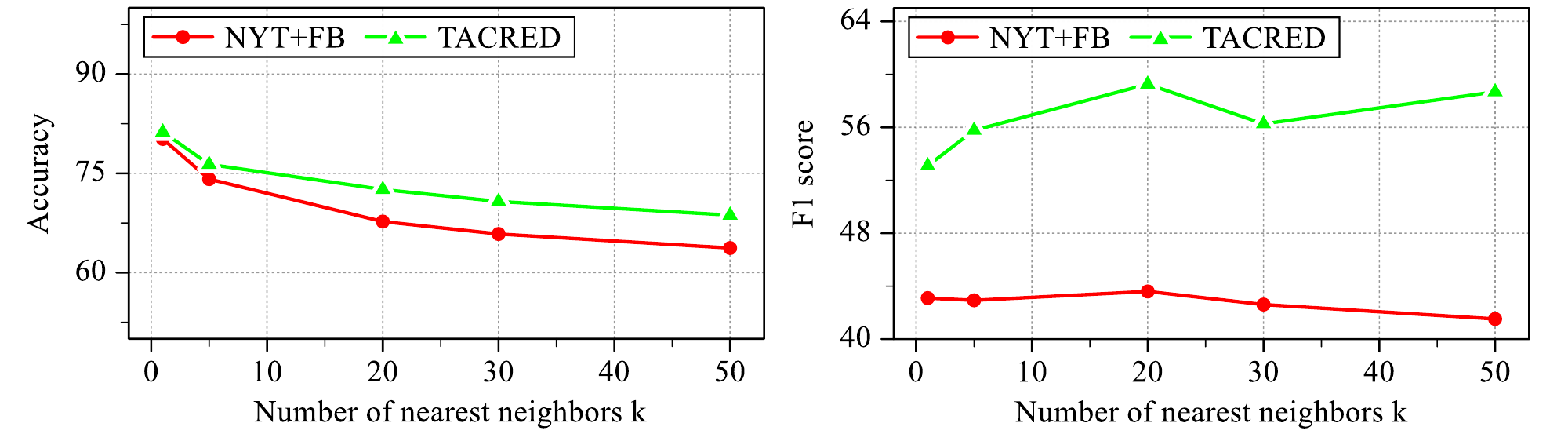}
    \caption{Influence of the used number of neighbors K.}
    \label{fig:number neighbors}
\end{figure}

\subsection{Results}
We summarize the performances of our method and seven baseline models in Table \ref{tab:result}. All of these models are evaluated on identical test set to show their performance. From the experimental results, we can see that our method significantly outperforms baselines. For NYT+FB dataset, compared with the previous SOTA model, our method improves V-measure F1 by 1.6\%, and ARI by 5.7\%, but the B$^{3}$ F1 score is only by 0.6\%. The reason why the performance gain is minuscule in NYT+FB is that the dataset contains numerous wrongly labeled instances in the train and test sets. These instances are unable to reflect the real performance of the model, and the number of samples in different relations is very unbalanced. As shown in Figure \ref{fig:dataset distribution}, the relation distribution is similar to a long-tailed distribution. Besides, most relations only have several samples that make a lot of noisy nearest neighbors to hurt performance when use mining nearest neighbors in Relation Semantic Clustering. For TACRED, compared with the previous SOTA model, our method improves B$^{3}$ F1 by 3.7\%, V-measure F1 by 7.0\%, and ARI by 10.1\%. It is worth noting that the score of precision and recall in B$^{3}$ seems to be extremely disequilibrium. By definition, precision measure the correct rate of putting each sample in its cluster and recall measure the correct rate of clustering all samples into a single class. Therefore, the results indicate that most of the samples from corresponding relations are clustered in the same cluster.

\noindent\textbf{Ablation Study.}\quad 
To study the contribution of different components in the proposed method, we conduct an ablation study on each component. For fair comparisons, the other settings remain the same as the main model. From Table \ref{tab:result}, we can see that in both NYT+FB and TACRED, the model's performance is degraded if any component is removed, indicating that both modules are important for the final model performance. 

\subsection{Detailed Analysis}
\noindent\textbf{Hyperparameter Analysis.}\quad
On account of the importance of two hyperparameters dropout rate $r$ and number of nearest neighbors $K$ which is in the encoder module and cluster module respectively, we conduct a detail analysis on them. Firstly, to further study the role of dropout rate in relation instance encoder for data augmentation, we try out different rates and report the performance of B$^{3}$ F1 on NYT+FB and TACRED. As shown in Table \ref{table:rate}, we observe that all the variants underperform the default dropout rate $r=0.1$ of Transformers \cite{NIPS2017_3f5ee243}. Using small dropout rate will introduce small divergence so that it is difficult for our model to learn discriminative representation, while large dropout rate will make more noise and prejudice similar semantic information. Secondly, we study the influence of number of nearest neighbors $K$ in  cluster module for mining nearest neighbors and report the accuracy and F1 score on two datasets. As shown in Figure \ref{fig:number neighbors}, the accuracy (left),which is the correct rate of that nearest neighbors with their corresponding samples are all come from same relation, is gradually decrease with the increase of $K$. However, the result of B$^{3}$ F1 score (right) is not very sensitive to the value and even remain perform well when increasing $K$ to 50, despite the increasing value will introduce more noise. This is beneficial, since we do not have to fine-tune the value on new raw text.

\begin{table}[t]
\renewcommand\arraystretch{1.3}
\caption{Effects of different dropout rates $r$ on the NYT+FB and TACRED development sets.}\label{table:rate}
\centering
\resizebox{0.9\linewidth}{!}{
\begin{tabular}{ccccccc}
    \hline
    Dataset\ / \ $r$    &0.01    &0.05    &0.1    &0.15   &0.2    &0.5\\ \hline
 NYT+FB & 41.3 & 42.1 & \textbf{44.3} & 44.1  & 43.8  & 41.5\\
 TACRED & 50.4 & 52.1 & \textbf{58.7} & 57.6  & 57.1  & 51.1\\\hline
\end{tabular}
}
\end{table}

\begin{figure}
    \centering
    \includegraphics[width=1\linewidth]{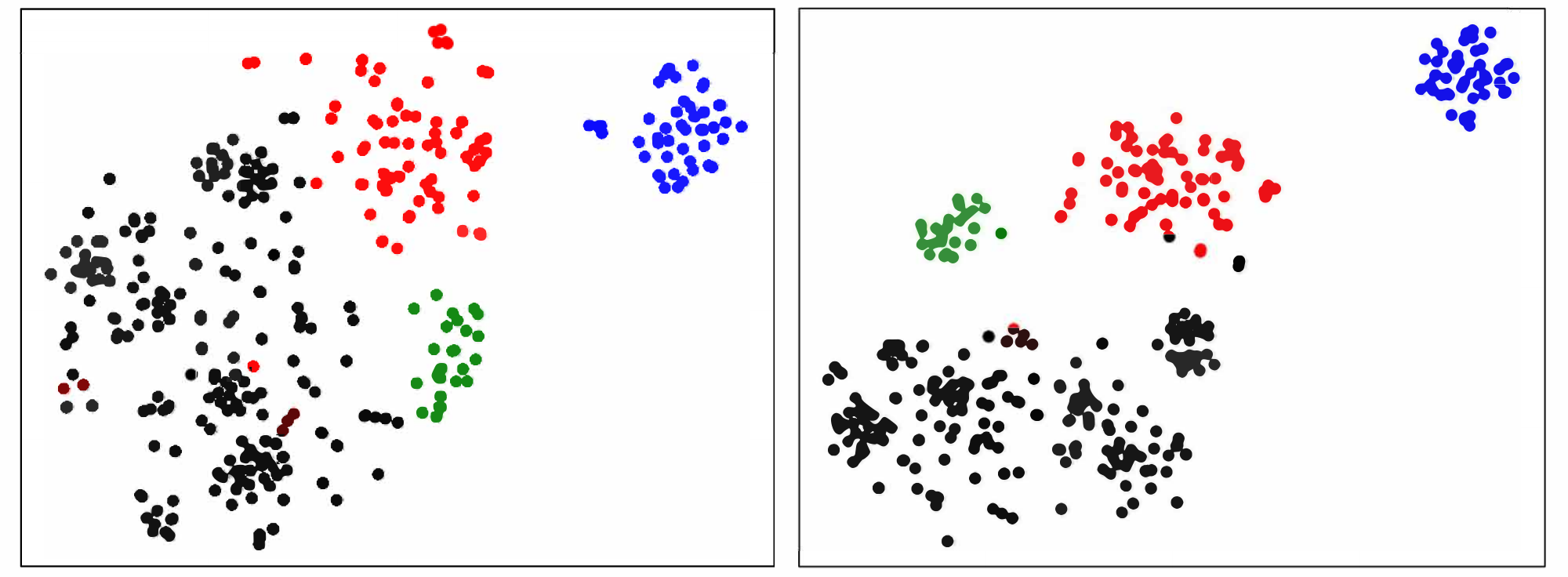}
    \caption{Visualizing contextualized entity-level features after t-SNE dimension reduction on TACRED.}
    \label{fig:visua}

\end{figure}

\noindent\textbf{Visualization of Relation Representations.}\quad 
In this experiment, to intuitively show the effectiveness of our model to learn representations in relational feature space, we visualize the representations of the instances in TACRED datasets with t-SNE \cite{van2008visualizing} and randomly select 4 relations from the test set. As shown in Figure \ref{fig:visua}, we color each instance according to its ground-truth relation label and we can observe that the proposed model without Relation Semantic Clustering (left) gives general results and does not provide discriminative cluster assignments. For example, the instances with black and red colors may have similar syntactic or surface features and clustering them directly will lead to a poor result. When we use clustering module in our model, model with full module (right) can learn more discrimintaive features and each relation is mostly separate from others.

\begin{figure}[t]
    \centering
    \includegraphics[width=1\linewidth]{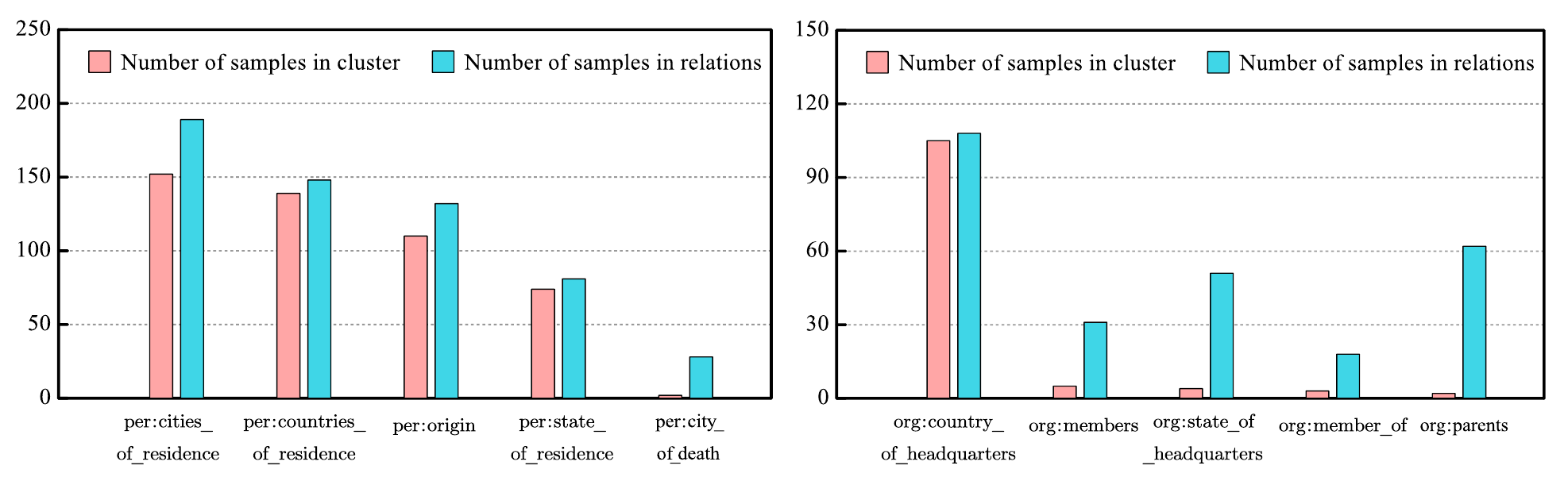}
    \caption{Statistics of samples in clusters and classes from the largest and smallest group.}
    \label{fig:statistic}
\end{figure}

\noindent\textbf{Analysis on Clustering Results.}\quad
Due to the number of clusters is lower than the number of true relations, different relations are likely to be clustered into same relation class. We attempt to have a detailed analysis on clustering results from TACRED to further verify whether different relation types in same cluster group have similar semantic or not. Specifically, we select the largest and smallest cluster group, the two most typical groups, which contain the most and least number of samples, to conduct detailed analysis. We find the top 5 real relations that appear most frequently in each of these two cluster groups. The 5 relation types in the former are: \textit{``per: cities\_of\_residence, per: countries\_of\_residence, per: origin, per: stateorprovinces\_of\_residence, per: city\_of\_death''}; The 5 relation types in the latter are: \textit{``org: country\_of\_headquarters, org: members, org: stateorprovince\_of\_headquarters, org: member\_of, org: parents}. The findings of the analysis have led to the conclusion that these relation types in the same cluster have analogous relational semantics and a potential hierarchical structure. 
Furthermore, we count the exact number of samples based on their true relation types in each cluster group.
As shown in Figure \ref{fig:statistic}, in the largest cluster group (left), the number of samples from different relation types is nearly and these samples occupy the vast majority of their corresponding relation types, which indicate that the cluster is a fine super-class. On the contrary, the most frequent relation type dominants the smallest cluster group (right), while other types only have one or few samples in this cluster, which indicate that the cluster is very pure and have less small impurities.

\section{Related Work}
\noindent\textbf{Self-supervised Learning}\quad
Self-supervised learning enables AI systems to learn from orders of magnitude more data, which is important to recognize and understand patterns of more subtle, less common representations of the world. Specifically, self-supervised learning tries to learn an encoder that extracts generic feature representations from unlabeled datasets. Early work focuses on solving different artificially designed pretext tasks  that does not require any supervision and can be easily constructed on the dataset, such as predicting neighbor words \cite{mikolov2013distributed}, generating neighbor sentences \cite{kiros2015skip} for textual data, and denoising \cite{vincent2008extracting}, colorization \cite{larsson2016learning}, adversarial generative models \cite{donahue2019large} for image data. Nevertheless, the feature representations are tailored to the specific pretext tasks with limited generalization.

Recent self-supervised learning algorithms mainly solve an instance discrimination task. In these algorithms, instance-wise contrastive learning with InfoNCE loss function \cite{oord2018representation} is prominent. Instance-CL treats each instance in the dataset and its augmentations as an independent pair and tries to pull together the representations within each pair while pushing apart different pairs . Consequently, different instances are well-separated in the learned feature space with local invariance being preserved for each instance. Although Instance-CL may implicitly group similar instances together, it pushes representations apart as long as they are from different original instances, regardless of their semantic similarities. Thereby, the implicit grouping effect of Instance-CL is less stable and more data-dependent, giving rise to worse representations in some cases \cite{li2020prototypical,purushwalkam2020demystifying}.

\noindent\textbf{Open Relation Extraction}\quad
Open relation extraction has received more attention in recent years and many efforts have been undertaken to exploring methods for it, due to the ability to extract new emerging relation types. The first line of research is Open Information Extraction , in which relation phrases are extracted directly to represent different relation types. However, using surface forms to represent relations results in an associated lack of generality since many surface forms can express the same relation. Another exploration is Relation Discovery, aims at discovering unseen relation types from open-domain text. Relation discovery can be divided into two different approaches: 1) cluster the relation representations learned from the instances, or 2) make more assumptions as learning signals to discover better relational representations. 

The variational autoencoder (VAE) based approaches are under unsupervised setting. Marcheggiani and Titov \cite{marcheggiani2016discrete} first propose the variational autoencoder method on unsupervised relation extraction. The model utilize the encoder extracts the semantic relation from hand-crafted features of the sentence and the decoder tries to predict one of the two entities given the relation and the other entity with a general triplet scoring function \cite{nickel2011three}. However, Simon et al. \cite{simon2019unsupervised} point out that the aforementioned method severely suffer from the instability, and they also propose two regularizers to guide the learning procedure. But the fundamental cause of the instability is still undiscovered. On this basis, Tran et al. \cite{tran-etal-2020-revisiting} demonstrate that by using only named entities to induce relation types can achieve better performance. Yuan et al. \cite{yuan-eldardiry-2021-unsupervised} assume that these classifications are a latent variable so they are required to follow a pre-defined prior distribution which results in unstable training and overcome this limitation by using the classifications as an intermediate variable instead of a latent variable.
In clustering-based approaches, the supervised learning model \cite{wu-etal-2019-open,zhang-etal-2021-open,zhao2021relation} are restricted by 
labeled data despite achieving good performance. In unsupervised setting,
Yao et al. \cite{yao2011structured} proposed Rel-LDA model ,using a generative model inspired by LDA to cluster sentences: each relation defines a distribution over a high-level handcrafted set of features describing the relationship between the two entities in the text (e.g. the dependency path). Hu et al. \cite{hu2020selfore} proposed SelfORE which encodes relational feature space in a self-supervised method that bootstraps relational feature signals by leveraging adaptive clustering and classification iteratively. 

\section{Conclusion}
In this work, we investigate the deficiencies of the contrastive learning on unsupervised relation extraction and propose a similarity-based representation learning method, which can learn well semantic of instances to effectively improve the performance with unsupervised clustering. In addition, we further obtain discriminative feature representations through relational semantic  clustering. Owing to the fact that our model is straightforward and efficient, we believe that our approach easily admits extensions to different open-domain texts. However, our model still has some shortcomings. 
Similar to other unsupervised relation extraction methods, our model is unable to handle instances where entity pairs appearing in a sentence do not exhibit any relation.
We leave these problems as future work and look forward to seeking possible solutions from a broader perspective.

\ack We would like to thank the referees for their comments, which
helped improve this paper considerably.

\bibliography{ecai}
\end{document}